\documentclass[sn-apa]{sn-jnl}

\usepackage{graphicx}%
\usepackage{multirow}%
\usepackage{amsmath,amssymb,amsfonts}%
\usepackage{amsthm}%
\usepackage{mathrsfs}%
\usepackage[title]{appendix}%
\usepackage{xcolor}%
\usepackage{textcomp}%
\usepackage{manyfoot}%
\usepackage{booktabs}%
\usepackage{algorithm}%
\usepackage{algorithmicx}%
\usepackage{algpseudocode}%
\usepackage{listings}%
\usepackage{appendix}
\usepackage[T1]{fontenc}

\theoremstyle{thmstyleone}%
\theoremstyle{thmstyletwo}%

\theoremstyle{thmstylethree}%

\newif\ifarxiv
\arxivtrue

\begin{document}

\title[Article Title]{Velocity Completion Task and Method for Event-based Player Positional Data in Soccer}

\ifarxiv
\author[1]{\fnm{Rikuhei} \sur{Umemoto}}\email{umemoto.rikuhei@g.sp.m.is.nagoya-u.ac.jp}

\author*[1,2]{\fnm{Keisuke} \sur{Fujii}}\email{fujii@i.nagoya-u.ac.jp}

\affil[1]{\orgdiv{Graduate School of Informatics}, \orgname{Nagoya University}, \orgaddress{\city{Nagoya}, \country{Japan}}}

\affil[2]{\orgdiv{Center for Advanced Intelligence Project}, \orgname{RIKEN}, \orgaddress{\city{Osaka}, \country{Japan}}}
\else
Anonymous
\fi 


\abstract{
In many real-world complex systems, the behavior can be observed as a collection of discrete events generated by multiple interacting agents. Analyzing the dynamics of these multi-agent systems, especially team sports, often relies on understanding the movement and interactions of individual agents. However, while providing valuable snapshots, event-based positional data typically lacks the continuous temporal information needed to directly calculate crucial properties such as velocity. This absence severely limits the depth of dynamic analysis, preventing a comprehensive understanding of individual agent behaviors and emergent team strategies. To address this challenge, we propose a new method to simultaneously complete the velocity of all agents using only the event-based positional data from team sports. Based on this completed velocity information, we investigate the applicability of existing team sports analysis and evaluation methods. Experiments using soccer event data demonstrate that neural network-based approaches outperformed rule-based methods regarding velocity completion error, considering the underlying temporal dependencies and graph structure of player-to-player or player-to-ball interaction. Moreover, the space evaluation results obtained using the completed velocity are closer to those derived from complete tracking data, highlighting our method's potential for enhanced team sports system analysis.
}


\keywords{machine learning, deep learning, information completion, missing data, sports, soccer}



\maketitle

\section{Introduction}
\label{sec:introduction}
Quantitative evaluation of teams and players has become essential for club officials and researchers as data analysis in team sports is increasingly important. Particularly, high-precision time-series data is required for improving gameplay and formulating effective opponent strategies. However, due to the complex movements of individual players based on their autonomous decision-making, team sports data suffers from low reproducibility and high acquisition costs. Hence, we can consider two approaches to address this issue: generating data through simulation or maximizing the utilization of existing data. 
The former approach of simulation design may benefit from combining two strategies \citep{fujii2023adaptive}. The first is an inverse approach, which aims to reproduce behavior from real-world data. The second is a forward approach, which generates behavior that maximizes rewards in a virtual environment without relying on observed data.
However, generating natural human-like behavior remains challenging, hindering practical applications. Therefore, this study focuses on the latter approach: maximizing the utilization of existing data.

Time-series data in team sports often contains missing values due to noise during acquisition, insufficient agent visibility in video footage, and partial non-disclosure of data items. Consequently, the importance of information completion techniques is increasing. Much of the existing research aims to approximate tracking data, which includes the positional information of all players and the ball at all times, but is costly to acquire. In sports such as soccer \citep{Omidshafiei2022,everett2023inferring,capellera2024transportmer,shan2023nrtsi}, basketball \citep{liu2019naomi,qi2020imitative}, and both of them \citep{choi2024dbhp,xu2024deciphering}, studies have attempted to complement the missing positional coordinates of the ball or players within specific time intervals using deep learning methods. However, as pointed out by \cite{penn2023continuous}, if the errors in the complemented positions are non-negligible, it can negatively impact subsequent player evaluations and team tactical analyses. Furthermore, many studies have addressed imputing positions from preceding and following information, given that some information is available. In other words, few studies have dealt with cases where the feature does not exist in the test datasets. Therefore, this study aims to complement information not in the test datasets when player positions are known.

One such data type is event data, including player coordinates at events such as passes and shots. This type of data enables more detailed analysis than previous event data, which contained only the coordinates of the on-the-ball player. Moreover, compared to tracking data, it is often publicly available. Therefore, it is expected to provide data analysis opportunities for fans and amateur clubs. Researchers also have begun to propose offensive \citep{rahimian2022let,robberechts2023xpass} and defensive \citep{umemoto2022location,umemoto2023evaluation} evaluation metrics using this data. However, since events such as passes and shots occur irregularly, event data is discrete concerning time. This makes it difficult to accurately calculate player velocities, which limits analytical applications such as space evaluation \citep{taki1996development,brefeld2019probabilistic,narizuka2021space,caetano2021football,spearman2018beyond,fernandez2018wide}. This means we can't fully understand individual player actions, how teams work together, or the physical reasons behind their movements.

Given the above background, this study proposes a task to complement player velocities, which do not exist in the test data, for analytical applications that require velocity information in team sports datasets. The proposed task takes only the positional coordinates of the ball and players at the time of events as input, and outputs player velocities. 
One of the benchmark tasks using player location and velocity data is to evaluate spaces in soccer, such as based on mathematical models \citep{spearman2018beyond}, which need to impute velocities. Therefore, this study aims to examine the effectiveness of our data-driven approach compared with the previous rule-based approach \citep{umemoto2023evaluation}.

The main contributions of this research are as follows:
\begin{enumerate}
    \item We present a new task to complete player velocity from player positional information only at the time of the event in team sports,
    \item We propose a new velocity completion method using a deep learning method, considering the graph structure and temporal information.
    \item Our experimental results demonstrate smaller errors with the true velocity and space evaluation method values closer to the actual data than the previous rule-based velocity completion \citep{umemoto2023evaluation}.
\end{enumerate}

\section{Related work}
\label{sec:related_work}
This section introduces existing research on information complements and evaluation indicators relevant to team sports for this paper.
    
\subsection{Information completion}
Information completion is not only the completion of missing values, but also the process of estimating and generating information not explicitly included in the data. Therefore, it is generally based on deep learning methods that can handle complex information and is used in situations that require more complex data processing, such as missing value completion for time series data \citep{che2018recurrent,cao2018brits,kidger2020neural,tashiro2021csdi} and missing region completion for image \citep{zhang2023image} and audio data \citep{ji2020comprehensive,morrone2021audio}. This information completion allows more information to be extracted from incomplete data, enabling analysis to gain a deeper understanding of complex phenomena in the real world.

Information completion for sports, a multi-agent time series data type, has recently attracted attention. The ultimate goal of this research is to reproduce tracking data, which includes the positional data of the ball and all players over time, but it is expensive to acquire. For example, there are studies on non-autoregressive models for complementing a single time point information \citep{liu2019naomi} or multiple time points \citep{shan2023nrtsi}, completion of player trajectories outside the broadcast video from past and future information \citep{Omidshafiei2022}, application of trajectory prediction models to past missing value completion \citep{qi2020imitative}, development of models capable of simultaneously handling trajectory prediction and completion \citep{xu2024deciphering, capellera2024transportmer}, and completion methods using velocity and acceleration information in addition to position information \citep{choi2024dbhp}. However, all of these have the issue of relying on tracking data for learning.

Research by \cite{penn2023continuous} and \cite{everett2023inferring} is similar to the problem setting of this research. Both predict player positions based on information about the limited number of players acquired at discrete intervals. The former uses a regression model and the Hungarian algorithm, while the latter employs a deep learning method that considers both graph structure and time series. This research differs from these two because it directly fills in the velocity not present in the test dataset.

\subsection{Evaluation for team sports}
Team sports have two main types of players and team evaluation metrics. One focuses only on players with the ball, and the other focuses on players who do not have the ball. In the former, the player evaluation is based on the event generated by the player who acted against the ball (See Subsection 3.4.2 of \cite{fujii2025machine}).

However, in many team sports, a player's non-ball possession time is much longer than their possession time, and considering this state is essential. Therefore, researchers have proposed evaluation metrics for the off-ball players' movements and space. In particular, the latter is crucial for considering a game's macro-level and micro-level characteristics. Space evaluation methods include those based on Voronoi diagrams \citep{taki1996development,fujimura2005geometric} or a bivariate normal distribution \citep{fernandez2018wide}, those considering player kinematic models \citep{brefeld2019probabilistic,martens2021space,caetano2021football}, and those weighting the field based on player time-to-arrival \citep{narizuka2021space}. Furthermore, \cite{spearman2018beyond} also proposed the Off-Ball Scoring Opportunities (OBSO), a probabilistic mathematical model with interpretability. This method has been extended to evaluate offensive soccer \citep{yeung2024strategic,teranishi2023evaluation}, basketball \citep{kono2024mathematical}, and ultimate \citep{iwashita2024space}. However, many of these studies rely on tracking data, which utilizes the velocity of all players and the ball at all times and is costly to acquire, thus limiting their application.

Recently, new data has been increasingly released in the soccer domain. This data is based on event data, including event types such as passes and shots, their success or failure, and ball coordinates at the start and end of events. It also contains player coordinates at the event occurrence. This enables more detailed analysis than traditional event data, and its public availability is expected to increase analysis opportunities for fans and amateur clubs. Consequently, researchers have proposed offensive \citep{rahimian2022let,robberechts2023xpass} and defensive \citep{umemoto2022location,umemoto2023evaluation} evaluation metrics using this data. In particular, \cite{umemoto2023evaluation} applied the  OBSO to this data and used it for defensive player positioning evaluation. However, the calculation of OBSO in that study uses rule-based velocities due to the discrete nature of the event data, which presents challenges to evaluation accuracy. This study aims to achieve a more valid evaluation by complementing these velocities with deep learning techniques.

\section{Methods}
\begin{figure}[]
    \centering
    \includegraphics[width=\textwidth]{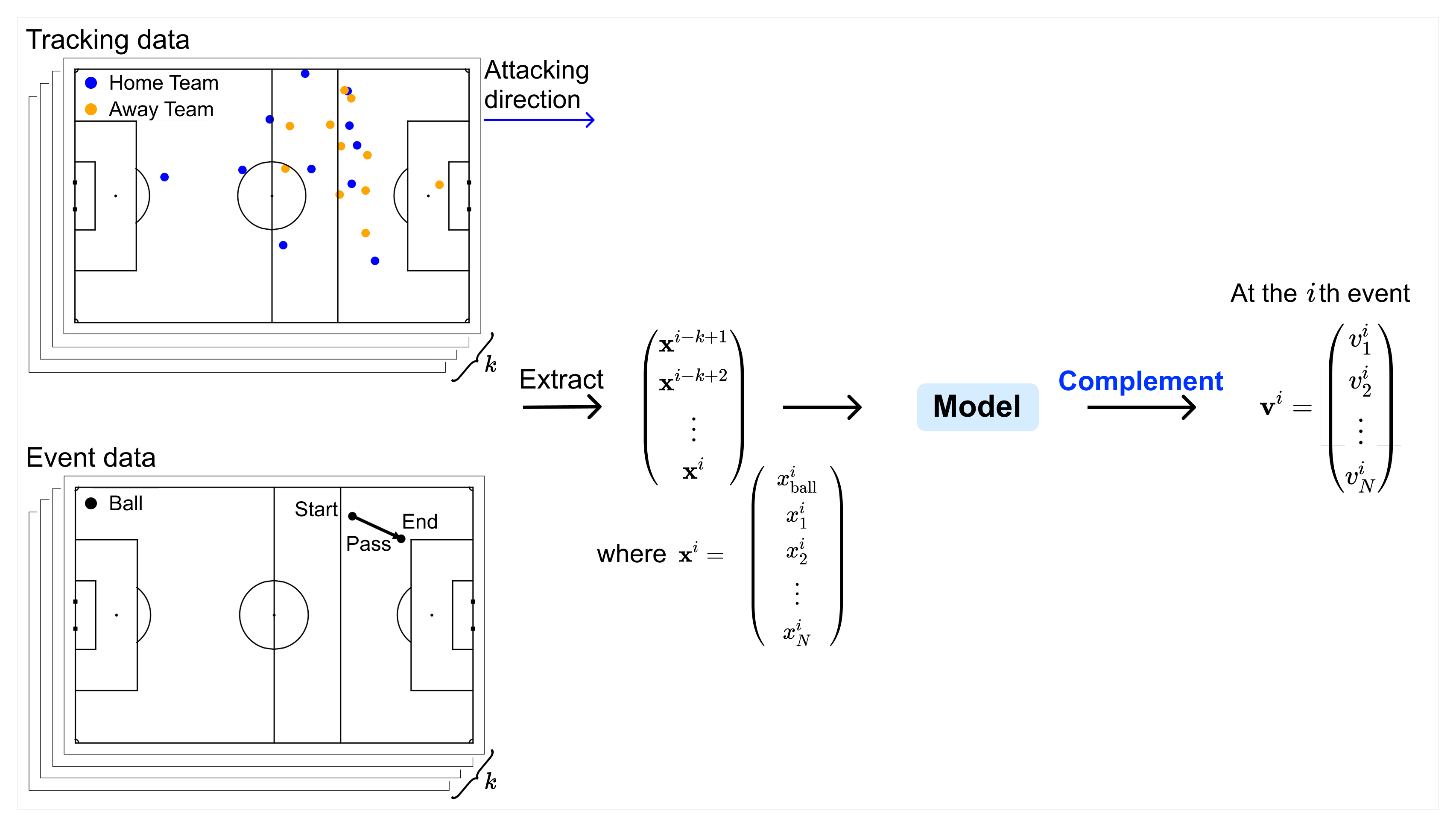}
    \caption{Flow chart of the task framework proposed in this study. We first extracted the coordinates of the ball and players from the tracking data and event data. We then input them into the model, and output the complemented player velocities.
    }
    \label{fig:task_framwork}
\end{figure}

This section describes the proposal for the definition of the player velocity completion task in this study and the model for this task. Fig. \ref{fig:task_framwork} depicts the details of this task.

\subsection{Definition of the Velocity Completion Task}
\label{ssec:def_task}
This study proposes a task to output player velocities only using ball and player positions collected at event times in football. Previous studies have addressed completing continuous player positions from discretely acquired positions \citep{penn2023continuous,everett2023inferring}. However, as pointed out by \cite{penn2023continuous}, large errors in their coordinates can lead to misinterpretations in analyses that require high data accuracy, such as space evaluation \citep{spearman2018beyond}. To address this, our study focuses on completing player velocities, assuming that player and ball positions are known. The data used in this task and the definition are described below.

First, we explain the data used in this task. The data is constructed based on event data, which includes event types such as shots and passes and the ball's positions at the start and end of each event. In addition to this information, we include the ball's velocity at the start of the event and the positions and velocities of all players at the beginning of the event for model learning and testing. Furthermore, to compare with rule-based velocity completion \citep{umemoto2023evaluation}, we limit the target events for this task to those occurring in the attacking third (the area on the opponent side of the field when the field is divided into three parts) and define this set of events as the dataset $\mathcal{D}$. To levarage models that can handle temporal information, we also create a dataset $\mathcal{D}^{*}$, which consists of event sequences from the $(i-k+1)$th event to the $i$th event, where $i$ is the target event ($k=10$ in this study).

Next, we define the task of this study mathematically. For a target event $i$, let $x^{i}_{\mathrm{ball}} \in \mathbb{R}^2$ denote the position of the ball at the begining of the event, and $x^{i}_{j}, v^{i}_{j} \in \mathbb{R}^2$ denote the position and velocity of a player $j$, respectively. Note that player $j$ shall only be judged if he is a teammate of another player, a goalkeeper, or the player in possession of the ball. In other words, we do not have information on player jersey numbers to distinguish the players. Hence, the task of this study is to complete all player velocities $\mathbf{v}^{i} = (v^{i}_{1}, v^{i}_{2}, \ldots, v^{i}_{N})$ using the positions $\mathbf{x}^{i} = (x^{i}_{\mathrm{ball}}, x^{i}_{1}, x^{i}_{2}, \ldots, x^{i}_{N})$, which comprises the ball and the $N$ players. For the case of velocity completion using temporal information, we use the set of event sequences from the $(i-k+1)$th event to the $i$th event. Thus, the task is to estimate $\mathbf{v}^{i}$ from the series of ball and player positions $(\mathbf{x}^{i-k+1}, \ldots, \mathbf{x}^{i})$.

The baseline for this task is rule-based velocity completion \citep{umemoto2023evaluation}. The data used in this research included positions for some players but lacked velocities. Therefore, this research set the rule-based velocities to calculate existing evaluation metrics \citep{spearman2018beyond}. We adopt this as a reasonable comparison target. The evaluation metric used in this task is the Root Mean Squared Error (RMSE) $\mathrm{[m/s]}$. 
Given the predicted player velocities $\tilde{\mathbf{v}}^{i}$, the RMSE for the dataset $\mathcal{D}$ and $\mathcal{D}^{*}$ to test the models is formulated as follows.
\begin{equation}
    \mathrm{RMSE}_{\mathcal{D}} = \frac{1}{|\mathcal{D}|} \sum_{\mathcal{D}} \sqrt{\frac{1}{N}\sum_{N}{|\tilde{\mathbf{v}}^{i}-\mathbf{v}^{i}|}^{2}}, \mathrm{RMSE}_{\mathcal{D}^{*}} = \frac{1}{|\mathcal{D}^{*}|} \sum_{\mathcal{D}^{*}} \sqrt{\frac{1}{N}\sum_{N}{|\tilde{\mathbf{v}}^{i}-\mathbf{v}^{i}|}^{2}},
    \label{eq:eval_metric}
\end{equation}
where $|\mathcal{D}|$ and $|\mathcal{D}^{*}|$ represent the numbers of the dataset $\mathcal{D}$ and $\mathcal{D}^{*}$. We also assume the test data sets to match concerning $\mathcal{D}$ and $\mathcal{D}^{*}$.

\subsection{Model to Complement Player Velocities}
\label{ssec:grnn}
\begin{figure}[]
    \centering
    \includegraphics[scale=0.1]{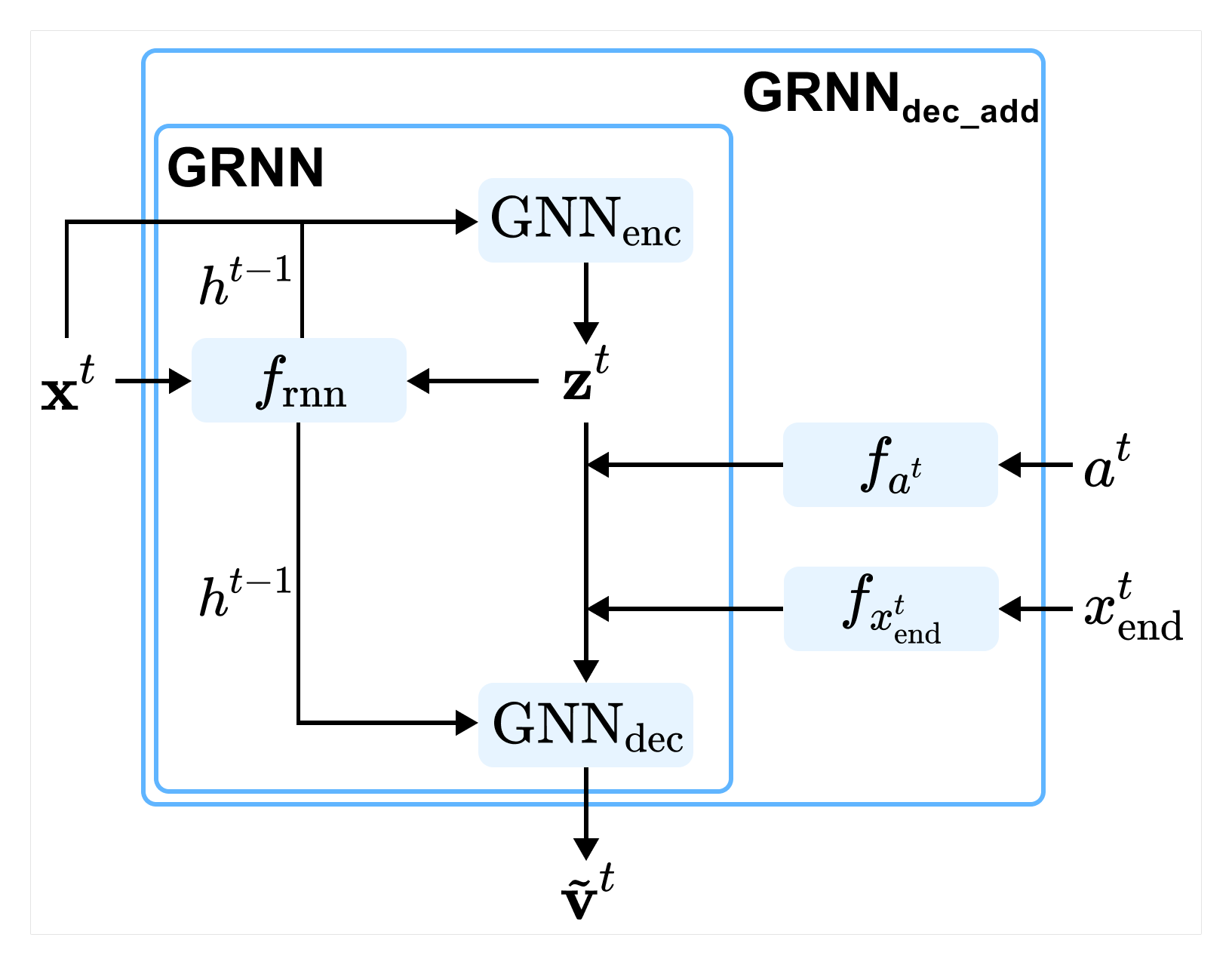}
    \caption{The GRNN model proposed in this study. If we also used the event action type ($a^{t}$) and ball coordinates at the end of the event ($x^{t}_{\mathrm{end}}$) as additional features, we selected $\mathrm{GRNN_{add}}$; otherwise, we utilized GRNN.
    }
    \label{fig:grnn}
\end{figure}

This section describes the models used for our task, as explained above. Fig. \ref{fig:grnn} shows the model's outline. 
Among related work in the previous section, we have found a study on deep learning-based methods for multi-agent information completion in team sports \citep{everett2023inferring}\footnote{\cite{everett2023inferring} focused on inferring off-ball player positions at the target event using features such as the positions of the ball or on-ball players around that event and the mean positions for each player in the game. In contrast, our task involves complementing player velocities at the target event using the positions of the ball and all players from several events preceding the target event.}.
Based on the study, we adopted a neural network that considered time series and graph structures called GRNN (Graph Recurrent Neural Network). Hence, after an overview of the GNN (Graph Neural Network), we describe the GRNN. 
Note that this architecture itself is not novel, but the novelty of our method is based on the new task (i.e., input and output modules of neural networks, and additional features. 
In the experiments, we will examine other network modules such as variational autoencoder and the combinations with RNN and GNN. 

\textbf{GNN:} one of the deep learning methods for dealing with graph data. This neural network method has become crucial in modeling and understanding complex relationships in the real world \citep{khemani2024review}. Therefore, researchers used this model for information completion methods in team sports \citep{Omidshafiei2022, everett2023inferring}.

In the following, we provide an overview of this study's GNN. Let the node features for ball or player $j$ be $o_{\mathrm{ball}}$ and $o_{j}$, and the edge features for player $j$ and ball or player $l$ be $e_{j,\mathrm{ball}}$ and $e_{jl}$, respectively at the $t$th event $(i-k+1 \leq t \leq i)$. We also defined the set of node features as $\mathbf{o} = (o_{\mathrm{ball}}, o_{1}, \ldots, o_{N})$. First, based on all the features of the nodes, we derived a function $f_{e}$ for all the features of the edge $\mathbf{e}$ between the nodes. During this node-to-edge transformation, we also added whether or not the player was a teammate and whether or not the player was a ball holder to the edge features. We then used a function $f_{o}$ that computed the edge features sent to each connection node to create new features for all nodes $\mathbf{o}^{\prime}$. Throughout this node-edge transformation process, we added whether it was a ball or not, whether it was a player or not, which was the direction of attack concerning the player, whether it was a ball holder or not, and whether it was a goalkeeper or not to the node features. We represented these operations as follows:
\begin{alignat}{2}
    \mathbf{e} &= f_{e}(\mathbf{o}), \label{eq:edge_update} \\  
    \mathbf{o}^{\prime} &= f_{o}(\mathbf{e}). \label{eq:node_update}
\end{alignat}
We defined the module considering these equations as $\mathrm{GNN}(\bullet)$, and we calculated the predicted player velocities $\tilde{\mathbf{v}}^{i}$ as follows:
\begin{equation}
    \tilde{\mathbf{v}}^{i} = \mathrm{GNN}(\mathbf{x}^{i})
    \label{eq:gnn}
\end{equation}

Additionally, regarding only models with this GNN module, we added other features concerning the event action type ($a^{i}$) and ball coordinates at the end of the event ($x^{i}_{\mathrm{end}} \in \mathbb{R}^{2}$). 
We used these features in the form of outputs, obtained by passing $a^{i}$ through a linear function $f_{a}$ and $x^{i}_{\mathrm{end}}$ through a linear function $f_{x_{\mathrm{end}}}$. Given $\mathrm{GNN_{add}}(\bullet)$ module as the GNN module with the features, we computed $\tilde{\mathbf{v}}^{i}$ as follows:
\begin{equation}
    \tilde{\mathbf{v}}^{i} = \mathrm{GNN_{add}}(\mathbf{x}^{i}, f_{a}(a^{i}), f_{x_{\mathrm{end}}}(x^{i}_{\mathrm{end}})).
    \label{eq:gnn_add}
\end{equation}

\textbf{RNN:} one of the deep learning methods for dealing with sequence data. This neural network method has become fundamental in information completion data in team sports, such as tracking data \citep{liu2019naomi, shan2023nrtsi, qi2020imitative, Omidshafiei2022, everett2023inferring, choi2024dbhp} or event data \citep{everett2023inferring}. In this study, we employed one of the RNN methods, GRU or Gated Recurrent Units \citep{cho2014learning}.

We explain the RNN in this study. We first created two modules composed of an MLP explained in Subsection \ref{sssec:model}: encoder, and decoder. We then defined the latent variables $\mathbf{z}^{t}$ and $h^{t}$ at $t$th event. The former was the output from the encoder's function $f_{\mathrm{enc}}$, and the latter was updated by the function $f_{\mathrm{rnn}}$ for the RNN. Given these definitions, we predicted the velocities from the following equations:
\begin{alignat}{3}
    \mathbf{z}^{t} &= f_{\mathrm{enc}}(\mathbf{x}^{t}, h^{t-1}), \label{eq:enc_rnn} \\
    \tilde{\mathbf{v}}^{t} &= f_{\mathrm{dec}}(\mathbf{z}^{t}, h^{t-1}), \label{eq:dec_rnn} \\
    h^{t} &= f_{\mathrm{rnn}} (\mathbf{x}^{t}, \mathbf{z}^{t}, h^{t-1}).
    \label{eq:update_rnn}
\end{alignat}

\textbf{GRNN (ours):} the proposed model based on the GNN and RNN. We first defined the encoder module, $\mathrm{GNN_{enc}}$, and the decoder one, $\mathrm{GNN_{dec}}$. Given the latent variables $\mathbf{z}^{t}$ for the outputs of $\mathrm{GNN_{enc}}$, and the latent variable $h^{t}$ for RNNs updated by the function $f_{\mathrm{rnn}}$, the predicted player velocities $\tilde{\mathbf{v}}^{t}$ were calculated from the following equations using equations \ref{eq:gnn}, \ref{eq:enc_rnn}-\ref{eq:update_rnn}:
\begin{alignat}{3}
    \mathbf{z}^{t} &= \mathrm{GNN}_{\mathrm{enc}}(\mathbf{x}^{t}, h^{t-1}), \\
    \tilde{\mathbf{v}}^{t} &= \mathrm{GNN}_{\mathrm{dec}}(\mathbf{z}^{t}, h^{t-1}), \label{eq:dec_grnn} \\
    h^{t} &= f_{\mathrm{rnn}} (\mathbf{x}^{t}, \mathbf{z}^{t}, h^{t-1}).
\end{alignat}
Additionally, in the case of the GRNN with the features $a^{t}$ and $x^{t}_{\mathrm{end}}$ ($\mathrm{GNN}_{\mathrm{dec\_add}}$), we calculated the velocity $\tilde{\mathbf{v}}^{t}$ using equations \ref{eq:gnn_add} and \ref{eq:dec_grnn} as follows:
\begin{equation}
    \tilde{\mathbf{v}}^{t} = \mathrm{GNN}_{\mathrm{dec\_add}}(\mathbf{z}^{t}, f_{a}(a^{t}), f_{x_{\mathrm{end}}}(x^{t}_{\mathrm{end}}), h^{t-1}).
\end{equation}


\section{Experiments and Results}
This section aims to validate models for complementing player velocities and to apply the velocities to a previous evaluation method.

\subsection{Dataset and Preprocessing}
The dataset used in this study comprises event and tracking data from 55 games of the 2019 Meiji J1 League season, a professional soccer league in Japan, provided by DataStadium Inc. The event data included what event occurred at a given time or frame, where the ball was on the pitch, and where it ended. In contrast, the tracking data recorded the coordinates of the ball and all players at a rate of 25 Hz. 

Using this dataset, we created the datasets $\mathcal{D}$ used to train the non-RNN-structured models and $\mathcal{D}^{*}$ to train the RNN-structured models for the task explained in Section \ref{ssec:def_task}. The reason for preparing two datasets here was that not only was the number of trainable series different for models with and without RNN structure, but the increased series length also increased the possibility of missing information necessary for model training. Table \ref{tbl:num_dataset} also shows that $\mathcal{D}^{*}$, the dataset for models with RNN structure, has fewer numbers than $\mathcal{D}$, the dataset for models without RNN structure. First, we added the positions and velocities of the players to the event data. We calculated the velocities from the tracking data at frames where the events occurred. We then extracted the events that occurred in the attacking third, and created a dataset with just that event and a $k$th event. Simultaneously, we removed the data with the ball position $x^{i}_{\mathrm{ball}}$ and the ball velocity $v^{i}_{\mathrm{ball}} \in \mathbb{R}^{2}$ missing. We operated these for each of the 55 games and split them into training data for the first 40 games, validation data for the next eight games, and test data for the last seven games. Finally, we aligned the number of test datasets for $\mathcal{D}$ and $\mathcal{D}^{*}$ and present the final numbers in Table \ref{tbl:num_dataset}.
\begin{table}[]
    \centering
    \begin{tabular}{ccc}
         Type & $\mathcal{D}$ for not RNN-structured models & $\mathcal{D}^{*}$ for RNN-structured models \\ \hline
         Training & 17,118 & 14,616 \\
         Validation & 3,726 & 3,226 \\
         Test & 2,958 & 2,958 \\
    \end{tabular}
    \caption{Total number of training, validation, and test data for data sets $\mathcal{D}$ and $\mathcal{D}^{*}$.}
    \label{tbl:num_dataset}
\end{table}

\subsection{Model Validation for Player Velocity Completion}
\subsubsection{Models} \label{sssec:model}
To compare our model, we first introduce several deep learning methods. We also explain the graph-structured models with the additional features $a^{t}$ and $x^{t}_{\mathrm{end}}$, as described in Section \ref{ssec:grnn}. We used Xavier initialization \citep{glorot2010understanding} to initialize all fully connected layers in these models. We described the model hyperparameters in Appendix \ref{sec:hyperparameters}.

\noindent\textbf{Rule-based baseline:} As we explained in Section \ref{ssec:def_task}, we adopted the rule-based player velocity settings used in \citep{umemoto2023evaluation}.

\noindent\textbf{MLP:} We created a three-layer linear function: we introduced an ELU function \citep{clevert2015fast}, one of the activation functions, and batch normalization between the first and second layers, and another ELU function between the second and third layers. 

\noindent\textbf{VAE (Variational AutoEncoder):} This model was proposed by \cite{kingma2013auto}. 
We created this model's prior, encoder, and decoder using the exact structure of the MLP. Given the latent variables $\mathbf{z}^{i}$ from the prior and encoder at $i$th event, we assumed the prior distribution (prior) $p_{\theta}(\mathbf{z}^{i}|\mathbf{x}^{i})$, the approximate posterior distribution (encoder) $q_{\phi}(\mathbf{z}^{i}|\mathbf{x}^{i},\mathbf{v}^{i},\mathbf{z}^{i})$, the likelihood function (decoder) $p_\theta(\tilde{\mathbf{v}}^{i}|\mathbf{x}^{i},\mathbf{z}^{i})$ followed different normal distributions. Behind each distribution's MLP, we added a linear function to output the mean and linear and softplus functions to output a standard deviation.

\noindent\textbf{GNN:} This model and $\mathbf{GNN_{add}}$ was explained in Section \ref{ssec:grnn}. 
Based on the MLP, we calculated the node-to-edge function $f_{e}$ after batch normalization and the edge-to-node function $f_{o}$ after the second ELU function. 

\noindent\textbf{VGNN:} This model's original was proposed by \cite{kipf2016variational}. In this study, we combined the GNN and VAE, calling this model VGNN. We also validated this model and the $\mathbf{VGNN_{dec\_add}}$, which was the GNN decoder of this model with the additional features. 

\noindent\textbf{RNN:} We explained this model in Section \ref{ssec:grnn}. We created the functions $f_\mathrm{enc}$ and $f_\mathrm{dec}$ with the same structure as the MLP. We also leveraged teacher forcing during training \citep{williams1989learning}. 

\noindent\textbf{VRNN:} This model's original was proposed by \cite{chung2015recurrent}. This study combined the VAE and RNN, calling this model VRNN. 
We assumed the prior distribution (prior) $p_{\theta}(\mathbf{z}^{t}|\mathbf{x}^{t})$, the approximate posterior distribution (encoder) $q_{\phi}(\mathbf{z}^{t}|\mathbf{x}^{t},\mathbf{v}^{t},\mathbf{z}^{t})$, the likelihood function (decoder) $p_\theta(\tilde{\mathbf{v}}^{t}|\mathbf{x}^{t},\mathbf{z}^{t})$ followed different normal distributions. Like the VAE, we added a linear function to output the mean and linear and softplus functions to output a standard deviation behind each distribution.

\noindent\textbf{GRNN (Ours):} We validated this model and the $\mathbf{GRNN_{dec\_add}}$. 

\noindent\textbf{GVRNN:} This model's original was proposed by \cite{yeh2019diverse}. This study combined the GNN, VAE, and RNN, calling this model GVRNN. Since this model was validated in the previous study to complement unseen player positions from tracking data \citep{Omidshafiei2022}, we adopted this model. In addition, we validated the $\mathbf{GVRNN_{dec\_add}}$, which was the GNN decoder of this model with the additional features. We explained the details of this model in Appendix \ref{sec:gvrnn}.

Next, we explain loss functions for the models we used. Due to differences in model structure and the number of datasets $|\mathcal{D}|$ and $|\mathcal{D}^{*}|$, we leveraged different functions for each model. 

\noindent\textbf{Non-RNN-structured models:} This means the MLP, GNN, $\mathrm{GNN_{add}}$, VAE, VGNN, and $\mathrm{VGNN_{dec\_add}}$ for the dataset $\mathcal{D}$. Using the definitions, we formulated the loss functions of these models, $\mathcal{L}_{\mathrm{MSE}\mathchar`-\mathcal{D}}$, as follows: 
\begin{equation}
    \mathcal{L}_{\mathrm{MSE}\mathchar`-\mathcal{D}} = \frac{1}{|\mathcal{D}|} \sum_{\mathcal{D}} \left[\frac{1}{N}\sum_{N}{|\tilde{\mathbf{v}}^{i}-\mathbf{v}^{i}|}^{2} - \mathcal{L}_{\mathrm{KL}\mathchar`-\mathcal{D}}\right],
\end{equation}
where $\mathcal{L}_{\mathrm{KL}\mathchar`-\mathcal{D}}$ represented the KL divergence for the VAE-structured models, and was calculated as follows:
\begin{equation}
    \mathcal{L}_{\mathrm{KL}\mathchar`-\mathcal{D}} = D_{\mathrm{KL}}(q_{\phi}(\mathbf{z}^{i}|\mathbf{x}^{i},\mathbf{v}^{i})||p_{\theta}(\mathbf{z}^{i}|\mathbf{x}^{i})).
\end{equation}
This term was zero for the non-VAE-structured models.

\noindent\textbf{RNN-structured models:} This means the RNN, GRNN, $\mathrm{GRNN_{dec\_add}}$, VRNN, GVRNN, and $\mathrm{GVRNN_{dec\_add}}$ for the dataset $\mathcal{D}^{*}$. Using the definitions, we formulated the loss functions of these models, $\mathcal{L}_{\mathrm{MSE}\mathchar`-\mathcal{D}^{*}}$, as follows: 
\begin{equation}
    \mathcal{L}_{\mathrm{MSE}\mathchar`-\mathcal{D}^{*}} = \frac{1}{|\mathcal{D}^{*}|} \sum_{\mathcal{D}^{*}} \frac{1}{k} \sum^{i}_{t=i-k+1} \left[\frac{1}{N} \sum_{N}{|\tilde{\mathbf{v}}^{t}-\mathbf{v}^{t}|}^{2} - \mathcal{L}_{\mathrm{KL}\mathchar`-\mathcal{D}^{*}}\right],
\end{equation}
where $\mathcal{L}_{\mathrm{KL}\mathchar`-\mathcal{D}^{*}}$ represented the KL divergence for the VAE-structured models, and was calculated as follows:
\begin{equation}
    \mathcal{L}_{\mathrm{KL}\mathchar`-\mathcal{D}^{*}} = D_{\mathrm{KL}}(q_\phi(\mathbf{z}^{\leq t}|\mathbf{x}^{\leq t},\mathbf{v}^{\leq t},\mathbf{z}^{< t})||p_\theta(\mathbf{z}^{\leq t}|\mathbf{x}^{\leq t},\mathbf{z}^{< t})).
\end{equation}
This term was zero for the non-VAE-structured models.

\subsubsection{Model Validation Result for Velocity Completion}
\label{ssec:models_and_framework_validation}
\begin{table}[]
    \centering
    \caption{Results of this study's player velocity completion task for each model.}
    \begin{tabular}{cc} 
    Model & RMSE [m/s] \\ \hline
    Rule-based \citep{umemoto2023evaluation} & 4.51 \\
    MLP  & $2.29$ \\
    GNN & $2.27$ \\
    VAE \citep{kingma2013auto} & $2.37$ \\
    VGNN \citep{kipf2016variational} & $2.31$ \\
    RNN \citep{cho2014learning} & $1.98$ \\
    VRNN \citep{chung2015recurrent} & $2.27$ \\
    \textbf{GRNN (ours)} & $\mathbf{1.90}$ \\
    GVRNN \citep{yeh2019diverse} & $2.21$ \\
    \end{tabular}
    \label{tbl:verify}
\end{table}
\begin{table}[]
    \centering
    \caption{Results of this study's player velocity completion task for each graph-structured model using additional features.}
    \begin{tabular}{cc} 
    Model & RMSE [m/s] \\ \hline
    Rule-based \citep{umemoto2023evaluation} & 4.51 \\
    $\mathrm{GNN_{add}}$ & $2.23$ \\
    $\mathrm{VGNN_{dec\_add}}$ & $2.24$ \\
    $\mathbf{GRNN_{dec\_add}}$ & $\mathbf{1.93}$ \\
    $\mathrm{GVRNN_{dec\_add}}$ & $2.15$ \\
    \end{tabular}
    \label{tbl:verify_add}
\end{table}

Tables \ref{tbl:verify} and \ref{tbl:verify_add} show the model validation results in our proposed velocity completion task. The latter table shows the results for the graph-structured models with additional features $a^{t}$ and $x^{t}_{\mathrm{end}}$. We obtained the average RMSE from 10 inferences in the test dataset. These tables demonstrate that deep learning methods for velocity completion achieved fewer errors than the rule-based method \citep{umemoto2023evaluation}. This indicates the effectiveness of deep learning methods for the task of this study. In particular, the GRNN reduces the error by 58\% compared to the rule-based method, as shown in Table \ref{tbl:verify}. Consistent with previous research on player position completion \citep{everett2023inferring}, this suggests considering the graph structure and temporal information was effective.

\begin{figure}[]
    \centering
    \includegraphics[scale=0.15]{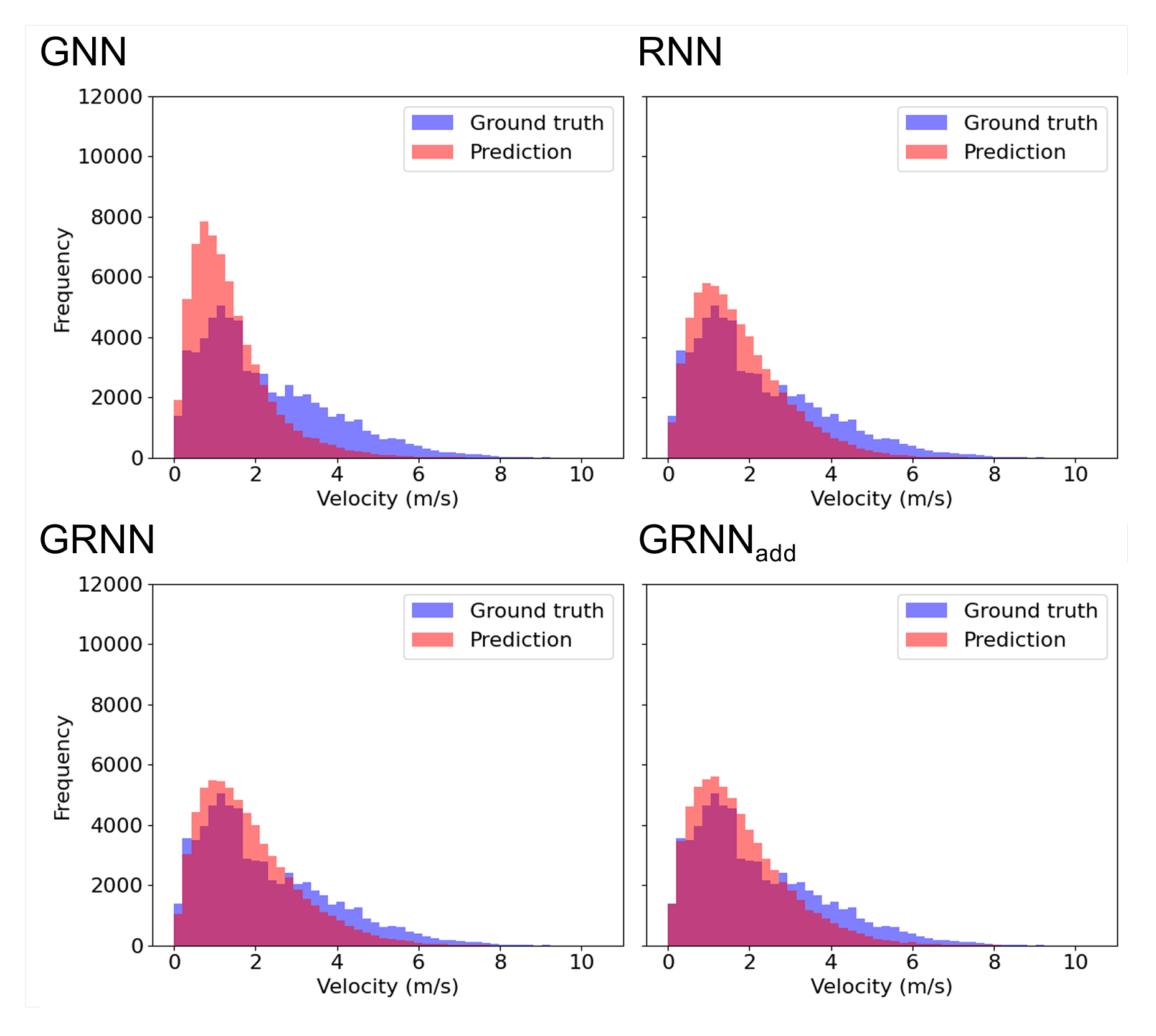}
    \caption{Results of comparing the distribution of predicted player velocities by GNN, RNN, $\mathrm{GRNN}$, and $\mathrm{GRNN_{dec\_add}}$ with the distribution of true velocities.}
    \label{fig:test_vel_distr}
\end{figure}

Next, we compared the predicted velocity distributions output by the models in the test dataset with the true velocity distributions. We show the results for GNN, RNN, $\mathrm{GRNN}$, and $\mathrm{GRNN_{dec\_add}}$ in Fig. \ref{fig:test_vel_distr}. The results for the other models are shown in Fig. \ref{fig:test_vel_distr_others} in Appendix \ref{sec:distr_other_models}. From Fig. \ref{fig:test_vel_distr}, it is clear that the RNN complemented velocity distributions closer to the true velocity distributions than the GNN. However, there is no difference between $\mathrm{GRNN}$ and $\mathrm{RNN}$ or $\mathrm{GRNN_{dec\_add}}$. Furthermore, it is difficult for any model to predict when the true velocities were high. One possible reason may be the small amount of data at higher velocities.

\subsection{Validation in the Application of the Complemented Player Velocities}
\subsubsection{Space Evaluation Metric}
We also compared our model against the rule-based baseline \citep{umemoto2023evaluation} using an existing space evaluation method to evaluate the applicability of the complemented velocities in the real world. We employed the Potential Pitch Control Field (PPCF) and Off-Ball Scoring Opportunities (OBSO) \citep{spearman2018beyond} as our evaluation metric in this study. We described the details of calculating these metrics in Appendix \ref{sec:ppcf}. The PPCF quantifies a player's ball control rate and necessitates both positional and velocity information for its computation. Consequently, previous studies have applied player position completion techniques within team sports data \citep{Omidshafiei2022, everett2023inferring}. Since our task focuses on velocity completion with known positions, specifically within the offensive-oriented attacking third, we also adopted OBSO, incorporating scoring probability with the PPCF. 

Using this framework, we defined the error metric $Er^{i}$ at the $i$th event to quantify the discrepancy between the complemented and true PPCF and OBSO values for each event we considered. Let $\mathrm{PPCF}^{i}_{r}$ and $\mathrm{OBSO}^{i}_{r}$ represent the ground-truth PPCF and OBSO value at a location $r$ on the field, and $\widetilde{\mathrm{PPCF}}^{i}_{r}$ and $\widetilde{\mathrm{OBSO}}^{i}_{r}$ denote the PPCF and OBSO value derived from the complemented velocities obtained through either the rule-based baseline or our proposed model, the GRNN, respectively. We then calculated the errors $Er^{i}_{\mathrm{ppcf}}$ about PPCF and $Er^{i}_{\mathrm{obso}}$ about OBSO as:
\begin{alignat}{2}
    Er^{i}_{\mathrm{ppcf}} &= \frac{1}{|\mathbb{R} \times \mathbb{R}|} \sum_{r \in \mathbb{R} \times \mathbb{R}} |\widetilde{\mathrm{PPCF}}^{i}_{r} - \mathrm{PPCF}^{i}_{r}|, \\
    Er^{i}_{\mathrm{obso}} &= \frac{1}{|\mathbb{R} \times \mathbb{R}|} \sum_{r \in \mathbb{R} \times \mathbb{R}} |\widetilde{\mathrm{OBSO}}^{i}_{r} - \mathrm{OBSO}^{i}_{r}|,
\end{alignat}
where $|\mathbb{R} \times \mathbb{R}|$ represented the total number of grid cells discretizing the playing field (in this study, $50 \times 32 = 1600$).

\subsubsection{Validation Result of Space Evaluation Metric}
\begin{figure}[]
    \centering
    \includegraphics[scale=0.10]{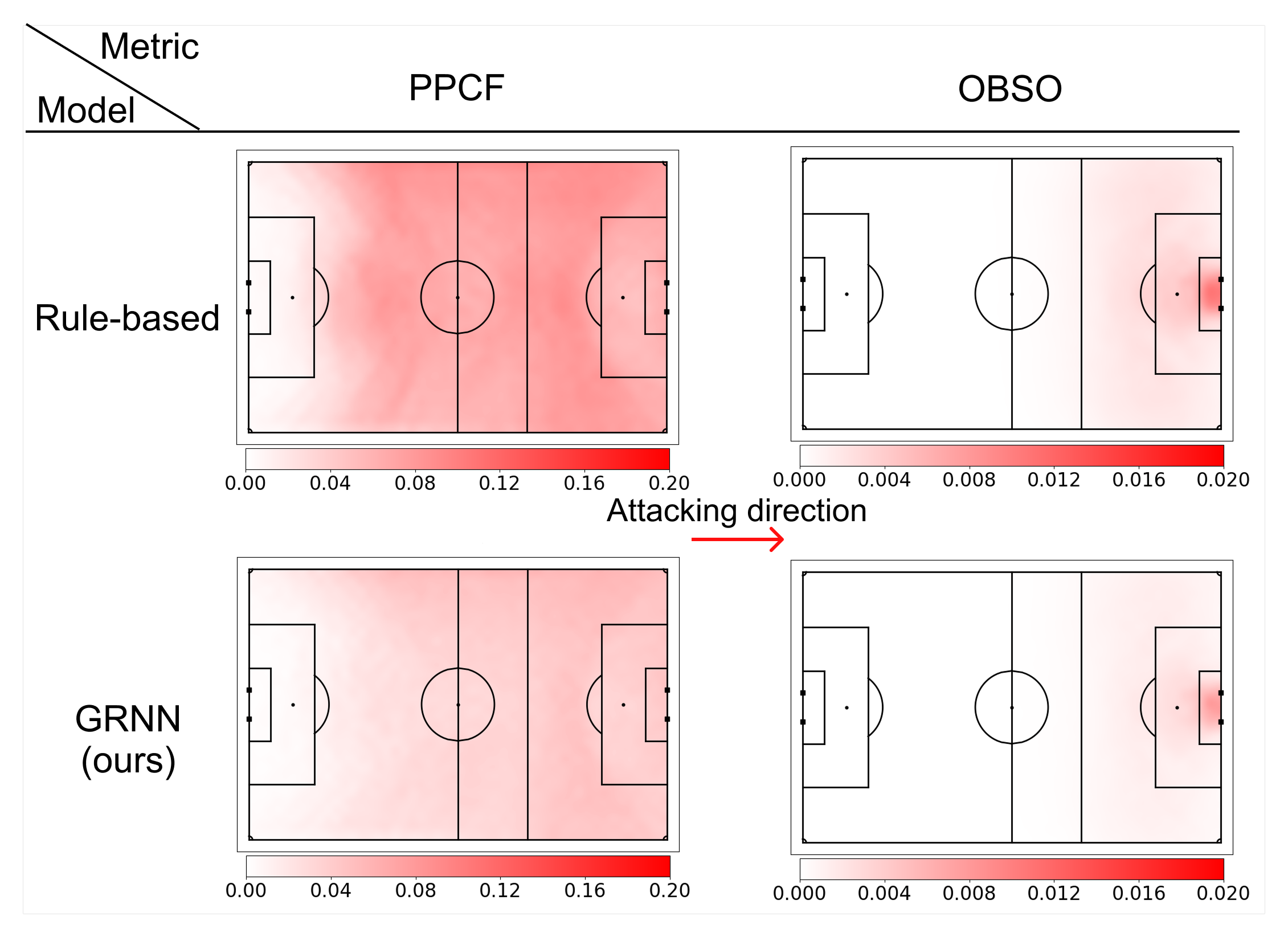}
    \caption{Error distribution over the pitch for PPCF and OBSO calculated using velocities complemented by the rule-based baseline or GRNN compared to ground truth.}
    \label{fig:error_distr}
\end{figure}
We first examined how the PPCF and OBSO values calculated using velocities complemented by either the rule-based baseline or the GRNN deviated from the ground truth across the entire pitch. The results are shown in Fig. \ref{fig:error_distr}. Darker red areas indicate larger errors against the ground truth in these visualizations. The figure shows that the GRNN leads to lighter red regions for PPCF and OBSO, indicating lower errors than the rule-based baseline. These findings suggest that the velocity completion model proposed in this study has greater applicability to existing evaluation metrics than rule-based velocity completion.

\begin{table}[]
    \centering
    \begin{tabular}{ccccccc}
     & \multicolumn{3}{l}{$Er^{i}_{\mathrm{ppcf}}$} & \multicolumn{3}{l}{$Er^{i}_{\mathrm{obso}}$} \\
     Model & Mean & Min & Max & Mean & Min & Max \\ \hline
     Rule-based \citep{umemoto2023evaluation} & 0.0570 & 0.0196 & 0.156 & 0.000817 & $8.62 \times 10^{-5}$ & 0.00594 \\
     GRNN (ours) & $\mathbf{0.0302}$ & $\mathbf{0.00431}$ & $\mathbf{0.128}$ & $\mathbf{0.000499}$ & $\mathbf{3.64 \times 10^{-5}}$ & $\mathbf{0.00367}$     
    \end{tabular}
    \caption{The errors $Er^{i}_{\mathrm{ppcf}}$ and $Er^{i}_{\mathrm{obso}}$ calculated using velocities complemented by the rule-based baseline or GRNN, and those were calculated using ground truth velocities. The Mean represented the error $Er^{i}_{\mathrm{ppcf}}$ or $Er^{i}_{\mathrm{obso}}$ averaged over the total number of test dataset.}
    \label{tbl:er_ppcf_obso}
\end{table}
\begin{table}[]
    \begin{tabular}{ccc|c}
     & Rule-based was smaller & GRNN was smaller & Total \\ \hline
     $Er_{\mathrm{ppcf}}$ & 99 & $\mathbf{2,859}$ & 2,958 \\
     $Er_{\mathrm{obso}}$ & 277 & $\mathbf{2,681}$ & 2,958
    \end{tabular}
    \caption{Table comparing the number of cases where the errors $Er_{\mathrm{ppcf}}$ and $Er_{\mathrm{obso}}$ were smaller for values calculated using rule-based velocities and GRNN-complemented velocities.}
    \label{tbl:num_better}
\end{table}
Next, Table \ref{tbl:er_ppcf_obso} presents the comparison results regarding the errors $Er_{\mathrm{ppcf}}$ and $Er_{\mathrm{obso}}$. This table also shows that PPCF and OBSO values calculated using GRNN-complemented velocities are closer to those computed with ground truth velocities. Furthermore, we counted the cases where the errors $Er_{\mathrm{ppcf}}$ and $Er_{\mathrm{obso}}$ for velocities complemented by either the rule-based method or GRNN were closer to the ground truth. The results are shown in Table \ref{tbl:num_better}. According to this table, the evaluation values obtained using GRNN-complemented velocities are closer to the ground truth more frequently for PPCF and OBSO. These results further support the conclusion that velocity complemented using our model leads to more realistic evaluations. As shown in Table \ref{tbl:verify}, the GRNN-complemented velocities exhibited smaller errors from the ground truth than those from the rule-based baseline, which was also reflected in the results of Fig. \ref{fig:error_distr}, and Tables \ref{tbl:er_ppcf_obso} and \ref{tbl:num_better}.

\begin{figure}[]
    \centering
    \includegraphics[scale=0.15]{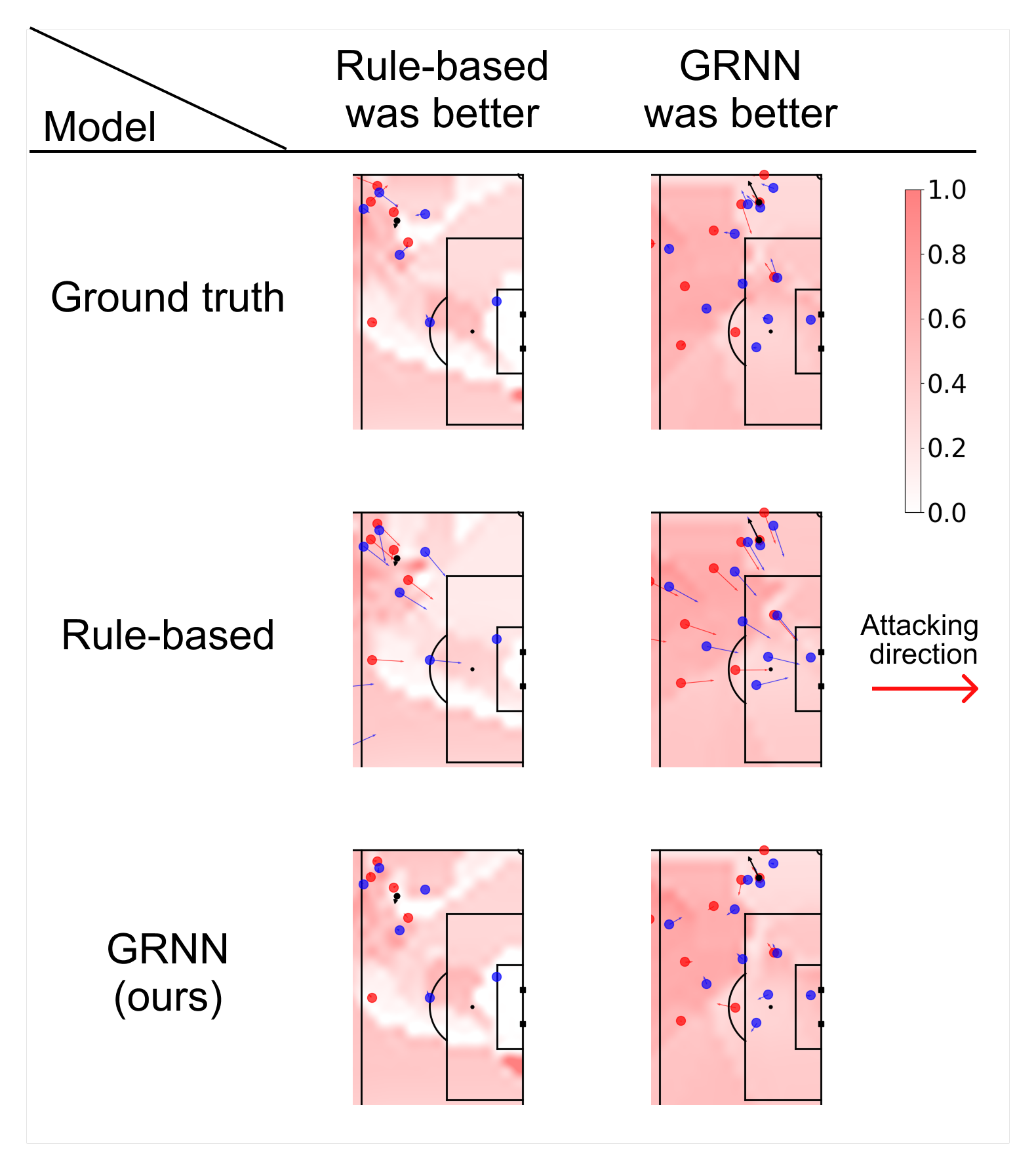}
    \caption{Example heatmaps of PPCF computed using ground truth, rule-based, or GRNN-complemented velocities.}
    \label{fig:compare_ppcf}
\end{figure}
\begin{figure}[]
    \centering
    \includegraphics[scale=0.15]{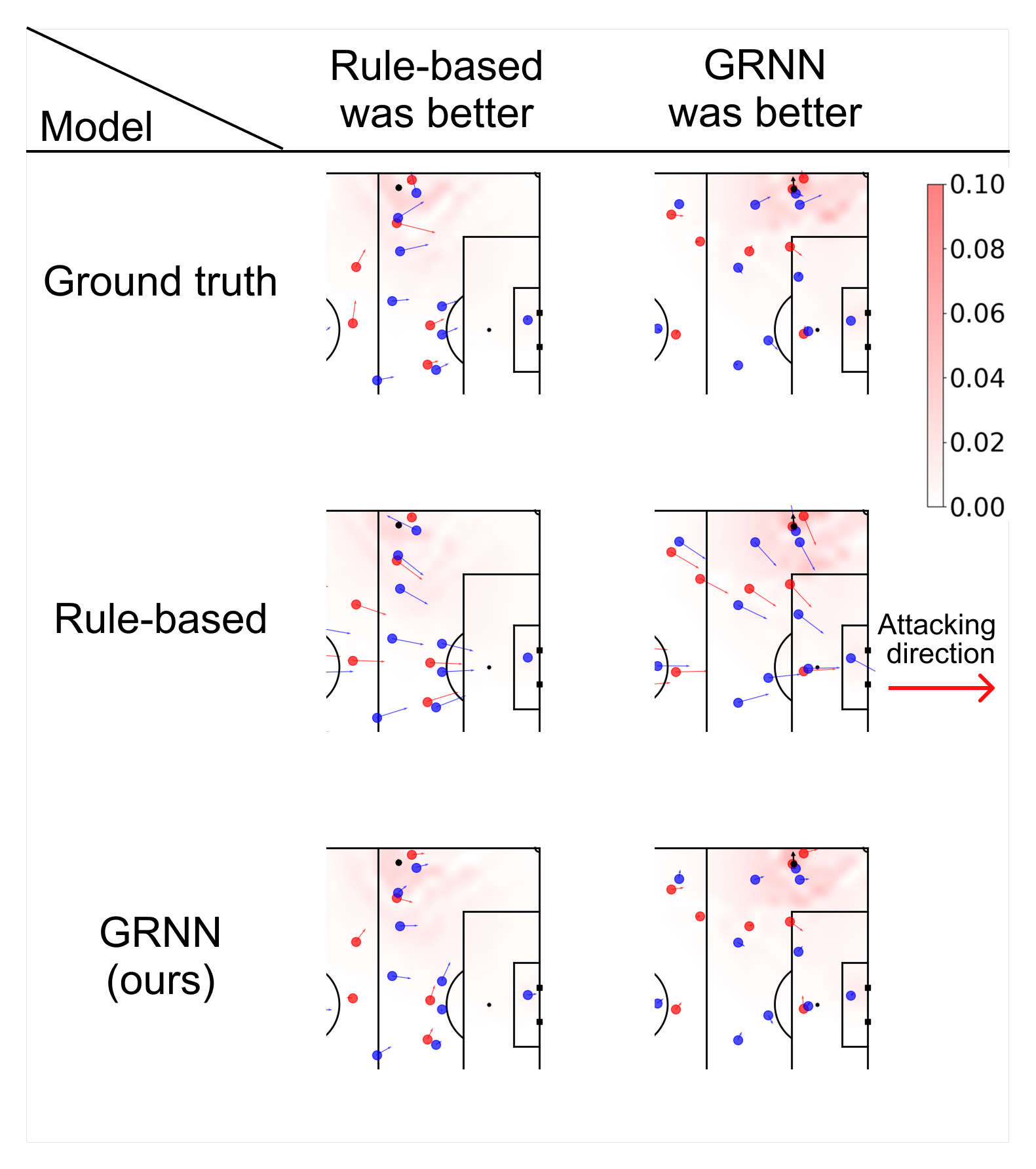}
    \caption{Example heatmaps of OBSO computed using ground truth, rule-based, or GRNN-complemented velocities.}
    \label{fig:compare_obso}
\end{figure}
Finally, Fig. \ref{fig:compare_ppcf} and \ref{fig:compare_obso} present examples of heatmaps of PPCF and OBSO computed using ground truth velocities and velocities complemented by the rule-based baseline and the GRNN. Darker red indicates higher PPCF or OBSO values for the attacking team in these heatmaps. For each figure, the left heatmap corresponds to the case where the result using the rule-based velocity was closer to the ground truth, while the right heatmap corresponds to the case where the GRNN result was closer. Qualitatively, there are no differences between the heatmaps, likely because we assumed player positions to be known. However, we can observe local differences caused by varying velocities. For example, in the top-right region of the ``GRNN was better'' heatmap in Fig. \ref{fig:compare_ppcf}, both the ground truth and GRNN results show lighter red, indicating lower attacking potential due to lower velocities of both attacking and defending players. In contrast, the rule-based result shows a darker red. Since this region is highly relevant for scoring opportunities in soccer, the GRNN-complemented result may be more appropriate, closer to the ground truth.

\section{Conclusion}
\label{sec:conclusion}
In this study, we propose a task to complement player velocities for data in which only the positions of all players at the time of an event and event information are available. In validating the model for this task, the GRNN, the deep learning method that considers the graph structure and time series information, minimized the error from the actual velocities. In addition, the velocities complemented by the GRNN produced PPCF and OBSO values, space evaluation metrics for soccer, closer to real data than those complemented using the rule-based baseline.

In the future, we expect to explore the following directions to improve and extend this study further. The first is the development of models with lower error. As shown in Table \ref{tbl:verify}, the GRNN achieved the lowest error value of 1.90 in this study. However, the error is still relatively large compared to previous sprint studies \citep{springham2020large, kavanagh2024relationships}. Therefore, we need to incorporate some models, such as Transformers \citep{vaswani2017attention}. The next is the expansion of velocity completion targets. This study defined a task to complete velocity only for events in the attacking third, to enable comparison with previous work \citep{umemoto2023evaluation}. Extending this task to cover all events is a potential direction. In addition, the data used in this study required the positional information of all players. However, many recently available datasets (e.g., StatsBomb\footnote{https://github.com/statsbomb/open-data}) do not include complete positional data, so methods for handling such incomplete data are also essential. Despite these challenges, we expect this framework to serve as a sports foundation and contribute to further data utilization and analysis.

\backmatter

\ifarxiv
\bmhead{Acknowledgments}
This work was financially supported by JST SPRING Grant Number JPMJSP2125 and JSPS KAKENHI Grant Number 23H03282. R. Umemoto would like to take this opportunity to thank the “THERS Make New Standards Program for the Next Generation Researchers”.
\fi

\section*{Declarations}
\begin{itemize}
\item Funding: This work was financially supported by JST SPRING  Grant Number JPMJSP2125 and JSPS KAKENHI Grant Number 23H03282.
\item Competing interests: The authors have no competing interests to declare that are relevant to the content of this article.
\item Ethics approval: Not applicable
\item Consent to participate: Not applicable
\item Consent for publication: Not applicable

\ifarxiv
\item Authors' contributions: All authors contributed to the study conception and design. Data preparation, modeling, and analysis were performed by Rikuhei Umemoto. The first draft of the manuscript was written by Rikuhei Umemoto and all authors commented on previous versions of the manuscript. All authors read and approved the final manuscript.
\end{itemize}
\else
\end{itemize}
\fi

\begin{appendices}
\setcounter{table}{5}
\setcounter{figure}{6}
\setcounter{equation}{18}

\section{GVRNN} \label{sec:gvrnn}
In this section, we explain one of the other models, GVRNN, using the definitions in Section \ref{ssec:grnn} and Subsection \ref{sssec:model}. The original GVRNN was proposed to predict the trajectory of each individual in a multi-agent environment \citep{yeh2019diverse}, which was the subject of comparison in the previous study \citep{Omidshafiei2022}. An example implementation of this model is available on GitHub, which we modified as a reference for our model
\ifarxiv
model\url{https://github.com/keisuke198619/C-OBSO}.
\else
model.
\fi
In this study, we called the GVRNN the combined models with the GNN, VAE, and RNN.

Given the definitions in Section \ref{ssec:grnn} and Subsection \ref{sssec:model}, predicted player velocities $\tilde{\mathbf{v}}^{t}$ was calculated from the following equations:
\begin{alignat}{4}
    p_\theta(\mathbf{z}^{\leq t}|\mathbf{x}^{\leq t},\mathbf{z}^{< t}) &= \mathcal{N}(\mathbf{z}^{t}|\boldsymbol{\mu}^{t}_{\mathrm{pri}},(\boldsymbol{\sigma}^{t}_{\mathrm{pri}})^2), \label{eq:vae_prior}\\   
    q_\phi(\mathbf{z}^{\leq t}|\mathbf{x}^{\leq t},\mathbf{v}^{\leq t},\mathbf{z}^{< t}) &= \mathcal{N}(\mathbf{z}^{t}|\boldsymbol{\mu}^{t}_{\mathrm{enc}},(\boldsymbol{\sigma}^{t}_{\mathrm{enc}})^2), \\
    p_\theta(\tilde{\mathbf{v}}^{t}|\mathbf{x}^{\leq t},\mathbf{z}^{\leq t}) &= \mathcal{N}(\tilde{\mathbf{v}}^{t}|\boldsymbol{\mu}^{t}_{\mathrm{dec}},(\boldsymbol{\sigma}^{t}_{\mathrm{dec}})^2) \label{eq:vae_decoder}, \\
    h^{t} &= f_{\mathrm{rnn}} (\mathbf{x}^{t}, \mathbf{z}^{t}, h^{t-1}),
\end{alignat}
where $\boldsymbol{\mu}^{t}_{\mathrm{pri}},\boldsymbol{\sigma}^{t}_{\mathrm{pri}}, \boldsymbol{\mu}^{t}_{\mathrm{enc}},\boldsymbol{\sigma}^{t}_{\mathrm{enc}}, \boldsymbol{\mu}^{t}_{\mathrm{dec}},\boldsymbol{\sigma}^{t}_{\mathrm{dec}}$ represented the mean and standard deviation of the prior, encoder, and decoder distributions. The output distributions of the prior, encoder, and decoder were also obtained using $\mathrm{GNN}(\bullet)$ as follows:
\begin{alignat}{3}
    [\boldsymbol{\mu}^{t}_{\mathrm{pri}},\boldsymbol{\sigma}^{t}_{\mathrm{pri}}] &= \mathrm{GNN}_{\mathrm{pri}}(\mathbf{x}^{t}, h^{t-1}), \\  
    [\boldsymbol{\mu}^{t}_{\mathrm{enc}},\boldsymbol{\sigma}^{t}_{\mathrm{enc}}] &= \mathrm{GNN}_{\mathrm{enc}}(\mathbf{x}^{t}, \mathbf{v}^{t}, h^{t-1}), \\
    [\boldsymbol{\mu}^{t}_{\mathrm{dec}},\boldsymbol{\sigma}^{t}_{\mathrm{dec}}] &= \mathrm{GNN}_{\mathrm{dec}}(\mathbf{z}^{t}, h^{t-1}).
\end{alignat}

\section{Hyperparmeters} \label{sec:hyperparameters}
This section describes the hyperparameters for the models to complement player velocities in this study. Table \ref{tbl:hp} showed the details.
\begin{table}[]
    \centering
    \begin{tabular}{cc}
    Hyperparameters & Value \\ \hline
    seed & 0 \\
    lr & $1.0 \times 10^{-4}$ \\
    num$\_$hid & 8 \\
    do$\_$prob & 0.0 \\
    rnn$\_$dim & 46 \\
    num$\_$layer & 1
    \end{tabular}
    \caption{Hyperparameters for the models used in this study.}
    \label{tbl:hp}
\end{table}
Where the explanation for each hyperparameter for the models is as follows:
\begin{itemize}
    \item seed: Fixed seed for machine learning.
    \item lr: The learning rate for machine learning.
    \item num$\_$hid: The dimensions of the latent variables.
    \item do$\_$prob: Dropout rate.
    \item rnn$\_$dim: The dimensions of the latent variables of the RNNs.
    \item num$\_$layer: Number of recurrent layers.
\end{itemize}

\section{OBSO and PPCF}
The Off-Ball Scoring Opportunities (OBSO) and the Potential Pitch Control Field were proposed by \cite{spearman2018beyond}. At location $r$ on the pitch in a current game state $D$, the former is formulated as follows:
\label{sec:ppcf}
\begin{equation}
P(G|D)=\sum_{r\in\mathbb{R}\times\mathbb{R}}P(G_r|C_r,T_r,D)P(C_r|T_r,D)P(T_r|D)
\end{equation}
where $T_r$ represents the probability of the next event that happens at $r$, $C_r$ represents the one of controlling the ball at $r$, and $G_r$ represents the one of scoring from $r$. \cite{spearman2018beyond} expressed $P(C_r|T_r,D)$ as the PPCF. This model is formulated as follows:
\begin{align}
\label{eq12}
\frac{dPPCF_j}{dT}(t,\overrightarrow{r},T|s,\lambda_j)&=(1-\sum_k PPCF_k(t,\overrightarrow{r},T|s,\lambda_j))f_j(t,\overrightarrow{r},T|s)\lambda_j \nonumber\\
f_j(t,\overrightarrow{r},T|s)&=\bigg[1+e^{-\pi\frac{T-\tau_{exp}(t,\overrightarrow{r})}{\sqrt{3}s}}\bigg]^{-1}
\end{align}
\noindent where $f_j(t,\overrightarrow{r},T|s)$ represents the probability that player $j$ at time $t$ will reach location $\overrightarrow{r}$ and control the ball within time $T$. $\lambda_j$ is the control rate per second for player $j$, $s$ is the temporal uncertainty on player-ball
intercept time and $\tau_{exp}(t,\overrightarrow{r})$ is the expected interception time based on the player's reaction time and maximum speed. In this study, we set the values of $\lambda_j$, $s$, reaction time, and maximum speed to the shared GitHub defaults\footnote{https://github.com/Friends-of-Tracking-Data-FoTD/LaurieOnTracking}. 

\section{The Velocity Distributions Calculated by The Other Models} \label{sec:distr_other_models}
\begin{figure}[]
    \centering
    \includegraphics[scale=0.15]{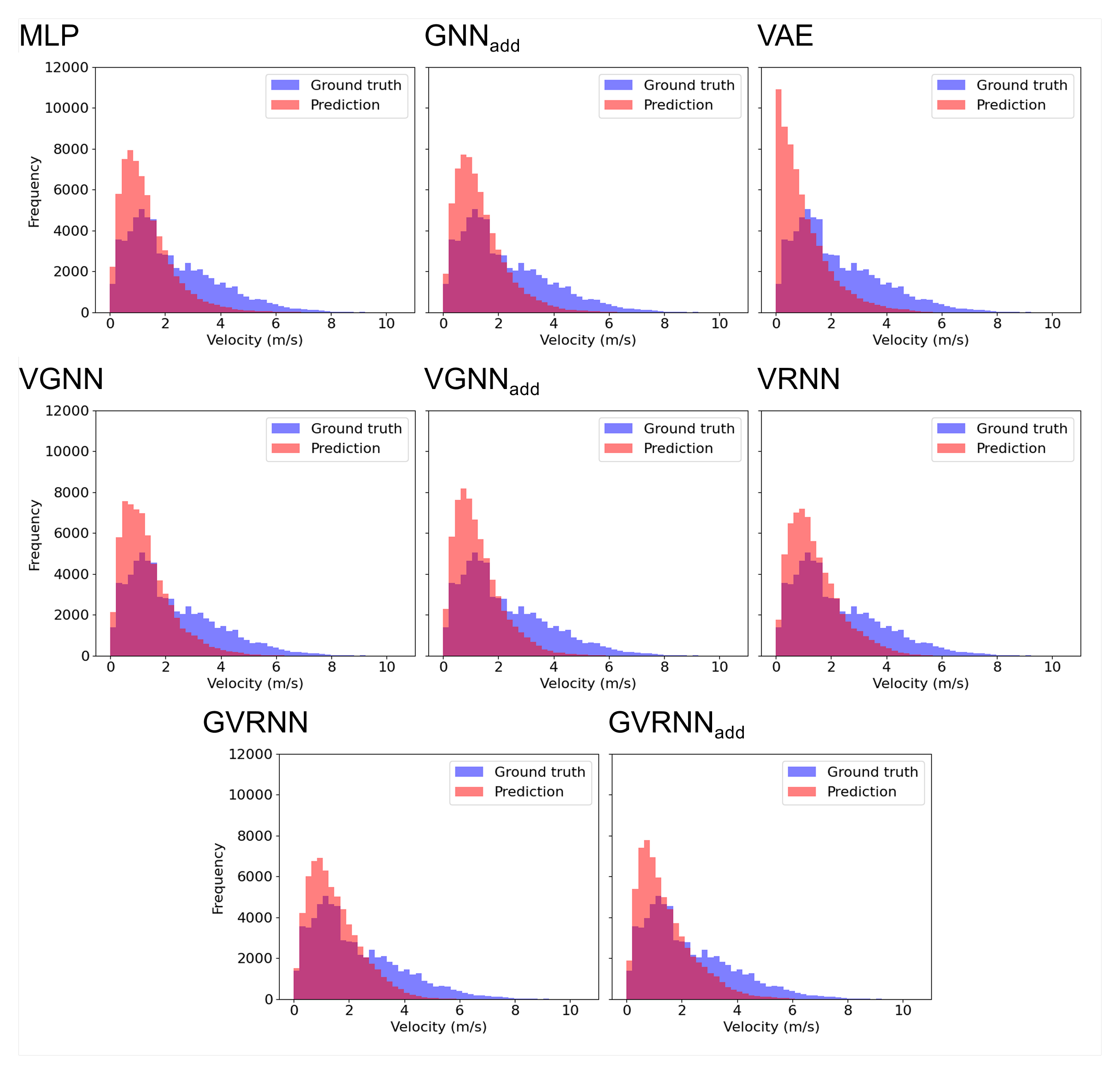}
    \caption{Results of comparing the distribution of predicted player velocities by MLP, $\mathrm{GNN_{add}}$, VAE, VGNN, $\mathrm{VGNN_{dec\_add}}$, VRNN, GVRNN, and $\mathrm{GVRNN_{dec\_add}}$ with the distribution of true velocities.}
    \label{fig:test_vel_distr_others}
\end{figure}
In Fig. \ref{fig:test_vel_distr_others}, the predicted velocity distributions from other models are compared with the true distributions in the test dataset. Tables \ref{tbl:verify} and \ref{tbl:verify_add} corroborated that the VAE-structured models poorly predicted player velocities, likely stemming from the assumption of a Gaussian distribution for latent variables and outputs. We can qualitatively see that the true velocity distribution (Fig. \ref{fig:test_vel_distr}) is not the Gaussian distribution. Football players do not frequently run fast, but walk and jog. This indicates the limitations of directly applying the previous VAE-based model with Gaussian assumptions for the player position completion task \citep{Omidshafiei2022}. In the future, we assume a skewed normal distribution and estimate its skewness as a parameter.

\end{appendices}

\newpage
\bibliography{sn-article}

\end{document}